\documentclass[conference]{IEEEtran}
\usepackage{cite}
\usepackage{amsmath,amssymb,amsfonts}
\usepackage{algorithmic}
\usepackage{graphicx}
\usepackage{textcomp}
\usepackage{xcolor}
\def\BibTeX{{\rm B\kern-.05em{\sc i\kern-.025em b}\kern-.08em
    T\kern-.1667em\lower.7ex\hbox{E}\kern-.125emX}}

\definecolor{wong-black}        {HTML}{000000}
\definecolor{wong-lightorange}  {HTML}{E69F00}
\definecolor{wong-lightblue}    {HTML}{56B4E9}
\definecolor{wong-green}        {HTML}{009E73}
\definecolor{wong-yellow}       {HTML}{F0E442}
\definecolor{wong-darkblue}     {HTML}{0072B2}
\definecolor{wong-darkorange}   {HTML}{D55E00}
\definecolor{wong-pink}         {HTML}{CC79A7}
\usepackage{csquotes}
\usepackage{booktabs}
\usepackage{hyperref}
\hypersetup{
    colorlinks=true,
    citecolor=wong-green,
    linkcolor=wong-darkblue,
    filecolor=wong-pink,      
    urlcolor=wong-black,
}
\usepackage[T1]{fontenc}
    
\begin{document}
\bstctlcite{IEEEexample:BSTcontrol}

\title{Label-Free Model Failure Detection\\for Lidar-based Point Cloud Segmentation}

\author{\IEEEauthorblockN{Daniel Bogdoll\IEEEauthorrefmark{2}\IEEEauthorrefmark{3}\textsuperscript{\textasteriskcentered},
    Finn Sartoris\IEEEauthorrefmark{2}\IEEEauthorrefmark{3}\textsuperscript{\textasteriskcentered},
    Vincent Geppert\IEEEauthorrefmark{2}\IEEEauthorrefmark{3}\textsuperscript{\textasteriskcentered},
    Svetlana Pavlitska\IEEEauthorrefmark{2}\IEEEauthorrefmark{3},
    and J. Marius Zöllner\IEEEauthorrefmark{2}\IEEEauthorrefmark{3}}

  \IEEEauthorblockA{\IEEEauthorrefmark{2}FZI Research Center for Information Technology, Germany\\
    bogdoll@fzi.de}
  \IEEEauthorblockA{\IEEEauthorrefmark{3}Karlsruhe Institute of Technology, Germany}}

\maketitle

\def\thefootnote{\textsuperscript{\textasteriskcentered}}\footnotetext{These authors contributed equally}\def\thefootnote{\arabic{footnote}}

\begin{abstract}
Autonomous vehicles drive millions of miles on the road each year. Under such circumstances, deployed machine learning models are prone to failure both in seemingly normal situations and in the presence of outliers. However, in the training phase, they are only evaluated on small validation and test sets, which are unable to reveal model failures due to their limited scenario coverage. While it is difficult and expensive to acquire large and representative labeled datasets for evaluation, large-scale unlabeled datasets are typically available. In this work, we introduce \textit{label-free model failure detection} for lidar-based point cloud segmentation, taking advantage of the abundance of unlabeled data available. We leverage different data characteristics by training a supervised and self-supervised stream for the same task to detect failure modes. We perform a large-scale qualitative analysis and present \textit{LidarCODA}, the first publicly available dataset with labeled anomalies in real-world lidar data, for an extensive quantitative analysis.
\end{abstract}

\section{Introduction}
\label{sec:introduction}

In machine learning, utilizing a fully labeled dataset for training, validation, and testing is typical. In autonomous driving, 70 - 85 \% of the data is usually reserved for training, leaving only 15 - 30 \% for both validation and testing~\cite{nuscenes, Sun_2020_CVPR, Yu2018BDD100KAD}. These small evaluation datasets stand in stark contrast to the millions of miles driven on public roads during deployment~\cite{forbes_60_million}. As a result, many failure modes of machine learning models, be it in seemingly normal situations or due to corner cases, are not captured in the evaluation sets. As large-scale unlabeled fleet data is generally available~\cite{mao2021million, Han2021SODA10MS, li2024multiagentmultitraversalmultimodalselfdriving}, there is an untapped potential to use this data for the detection of failure modes of machine learning models.

\begin{figure}[t]
  \centering
  \includegraphics[width=1\columnwidth]{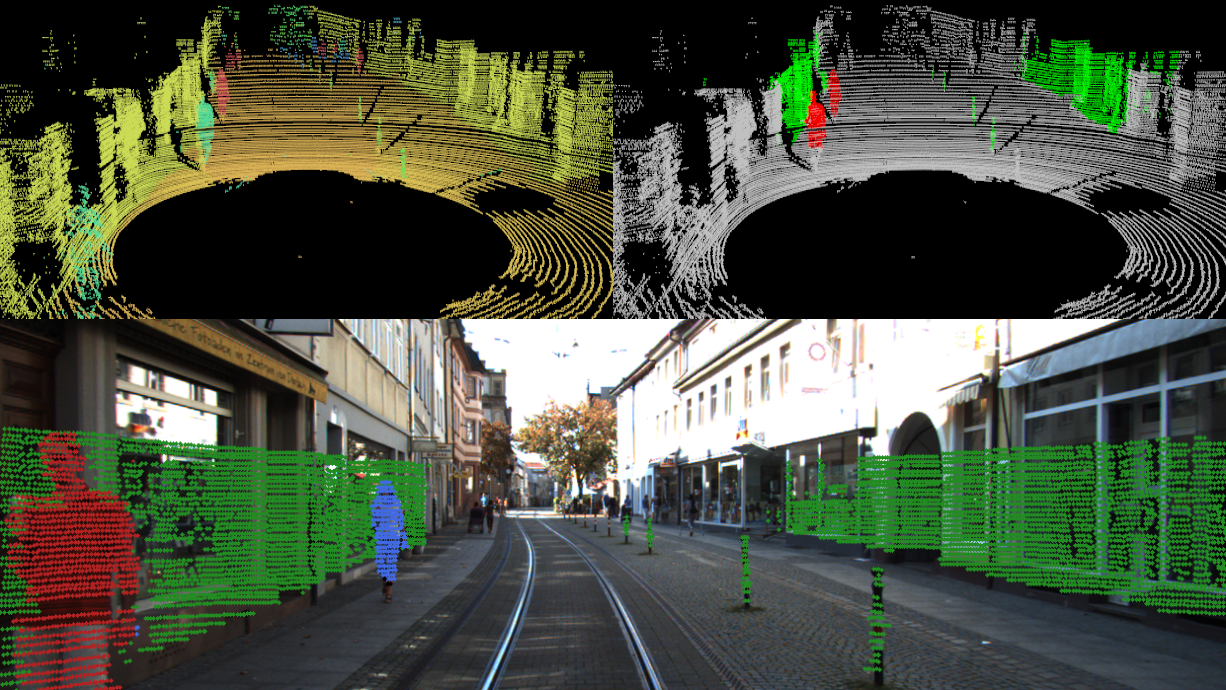}
   \caption{\textbf{Model Failure Detection.} The top left point cloud shows a \textit{supervised} and the top right a \textit{self-supervised} motion segmentation. The supervised model falsely classifies the pedestrian in the front left as static. Our approach exposes this model failure, as highlighted in red in the bottom image. More details on the color scheme can be found in Sec.~\ref{subsec:discrepancy}.}
   \label{fig:front_pic}
\end{figure}

There are many active research areas dealing with the detection of failure modes. Active learning~\cite{kumarActiveLearningQuery2020} is concerned with continuously enriching training data by querying samples from a set of unlabeled data points. Discrepancies between different sensor systems can also be used to query samples~\cite{wadatcvprCVPR21WAD2021}. In error estimation, many approaches try to utilize unlabeled test sets to evaluate models~\cite{dengAreLabelsAlways2021}. Label refinement compares given labels, e.g., by an auto-labeling process, with new proposals~\cite{schubertIdentifyingLabelErrors2024}. All of these methods have in common that they utilize or compare two or more different results for the same task. However, we are unaware of approaches that take advantage of different training paradigms to detect model failures. In this work, we introduce the concept of complementary learning to leverage different data characteristics of the training dataset, as shown in Fig.~\ref{fig:front_pic}. We train two models, one supervised and the other self-supervised, to detect model failures. Our main contributions are:

\begin{itemize}
\item Label-free model failure detection for lidar-based point cloud segmentation based on \textit{complementary learning} with a supervised and a self-supervised approach
\item \textit{LidarCODA}, the first publicly available real-world anomaly dataset with labeled lidar data, to quantitatively evaluate the sensitivity of the approach to outliers
\end{itemize}

All code to run our experiments and generate the LidarCODA dataset is available on \href{https://github.com/daniel-bogdoll/model_contradictions}{GitHub}\footnote{\href{https://github.com/daniel-bogdoll/model_contradictions}{github.com/daniel-bogdoll/model\_contradictions}}.
\section{Related Work}
\label{sec:related_work}

\begin{figure*}[t!]
  \centering
  \includegraphics[width=\textwidth]{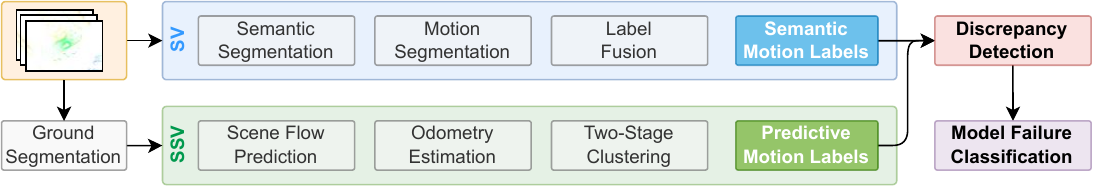}
  \caption{\textbf{Overview.} Given point clouds, we derive semantic motion labels in a supervised~(sv) fashion. In addition, we perform ground segmentation and derive predictive motion labels in a self-supervised~(ssv) fashion. Subsequently, we perform point-wise discrepancy detection and classify potential model failures.}
  \label{fig:method_overview}
\end{figure*}

The concept of comparing the outputs of two or more neural networks was already introduced in 1994 by Cohn et al., where they queried samples for active learning based on the disagreement between neural networks~\cite{cohnImprovingGeneralizationActive1994}. Since then, the variability in model predictions has been widely used to detect anomalies or errors. Ensemble diversity is especially well studied, as it was shown to lead to better performance~\cite{dietterich2000ensemble}, robustness~\cite{pang2019improving}, uncertainty quantification~\cite{lakshminarayanan2017simple}, and detection of outliers or distribution shifts~\cite{mehrtens2022improving, pagliardiniAGREEDISAGREEDIVERSITY2023, wangLearningDisagreementEvent2022}. While no uniform metric for ensemble diversity exists, measures like disagreement of models, double fault measure, or output correlation are widely used~\cite{kuncheva2003measures}. Ensemble diversity can be implicitly enhanced via random initialization~\cite{lakshminarayanan2017simple}, noise injection or dropout, or explicitly via bagging, boosting, or stacking. Compared to ensembles, mixtures of experts~\cite{jacobs1991adaptive} enforce higher model specialization and thus more component diversity, leading to better detection of out-of-distribution data~\cite{pavlitskayaUsingMixtureExpert2020a,pavlitskaya2022evaluating}. 

The approaches mentioned above involve a combination of several neural networks with similar or identical architectures. Active learning is another research field interested in the detection of model failures. Here, uncertainty derived from ensembles is resource-intensive and thus only rarely used as part of a querying strategy~\cite{schroderRevisitingUncertaintybasedQuery2022,roitberg2018informed,yahaya2019consensus}. Similar to ensembles, disagreements in a query-by-committee setting can be used to select samples~\cite{hinoActiveLearningQuery2022}. In autonomous driving, also contradicting detections from sensors can be used as triggers, e.g., when radar and camera detections do not match~\cite{wadatcvprCVPR21WAD2021}. Discrepancies between teacher and student models, typically known from knowledge distillation, can also be utilized~\cite{chenLearningEfficientObject2017a}. As test sets are often small and not representative, directly estimating the accuracy of a model with only unlabeled data is of high interest~\cite{dengAreLabelsAlways2021,pengEnergybasedAutomatedModel2024,baekAgreementontheLinePredictingPerformance,chenDetectingErrorsEstimating2021}. Here, we often see simple classification tasks or approaches that estimate an overall error that cannot be applied to individual samples. In some cases, generated pseudo-labels are utilized for further training steps~\cite{wangLearningFromDisagreementModelComparison2023,yuPredictingOutofDistributionError2022}.

Disagreements can also be used for detecting erroneous labels. Ground truth labels in large vision datasets are often error-prone when auto-labeling processes based on large models are employed~\cite{chachulaCombatingNoisyLabels2023}. Detecting label errors with disagreements can be done by predicting a novel or refined label, and uncertainties can be generated by predicting multiple such labels~\cite{huProbabilityDifferentialBasedClass2022,schubertIdentifyingLabelErrors2024,barNovelBenchmarkRefinement2023}. This way, also noisy labels introduced by human errors can be detected~\cite{northcuttConfidentLearningEstimating2021}.

Robustness during deployment is often achieved with sensor fusion, which, quite differently, purposefully aims to complement the weaknesses of one sensor with the strengths of another. Thus, disagreements are both typical and wanted, with the aim of resolving them~\cite{kowolYOdarUncertaintybasedSensor2020}. However, also data from a single sensor can be split into multiple streams to increase robustness. For example, object detection can be improved by combining appearance and geometry~\cite{intelnewsroomCES2021Hood2021} or temporality and geometry~\cite{Baur2024ECCV,lentsch2024union}. In performance monitoring~\cite{shaoWhenItLikely2024,buerkleSafePerceptionHierarchical2022}, but also in outlier or anomaly detection~\cite{delicOutlierDetectionEnsembling2024c}, typically, a primary model performing a regular task is accompanied by a learned or model-based module that provides some sort of uncertainty for the results of the regular task.

\textbf{Research Gap.} Many of the analyzed works utilizing disagreements deal with toy problems and only analyze classification tasks, which are not sufficient to truly understand the shortcomings of a model that is designed for the complex task of autonomous driving. Many works analyze model outputs of the same architecture, leveraging differences during training. However, this way, the same data characteristics are being used during training. Disagreement-based approaches in designing triggers for active learning~\cite{wadatcvprCVPR21WAD2021} and increasing robustness during deployment~\cite{intelnewsroomCES2021Hood2021} are most similar to our approach, but these industry demonstrations are not accompanied by scientific works and are thus hard to evaluate. Finally, to the best of our knowledge, no work exists that utilizes different training paradigms to detect model failures through disagreements.

\section{Method}
\label{sec:method}

To detect model failures without labeled validation or test sets, we perform complementary learning for the same task in order to detect model failures and classify challenging scenarios. We introduce the term \textit{complementary learning} for the use of different training methods, e.g., supervised and self-supervised, which complement each other for a given purpose, e.g., for the detection of model errors based on model predictions. The ability to detect model failures is based on the intuition that different training paradigms leverage different data characteristics from the same training dataset. We demonstrate this approach with the segmentation of lidar point clouds for autonomous driving into dynamic and static points, which we call motion labels. As shown in Fig.~\ref{fig:method_overview}, we first derive motion labels in a \textit{supervised} and \textit{self-supervised} fashion. Here, the first paradigm leverages human knowledge through labels, given only context from static scenes. On the other hand, the second paradigm leverages temporal information inherent in the data. Typically, these paradigms are combined either in a pre-training context~\cite{chenAdversarialRobustnessSelfSupervised2020} or with a combined loss during learning~\cite{chengSelfsupervisedSupervisedJoint2021}. Based on a point-wise comparison, we detect discrepancies and cluster them for better interpretation. Finally, an oracle examines and classifies the model failures to better understand challenging situations. 


\subsection{Supervised Semantic Motion Labels}
\label{subsec:supervised}

\begin{figure}[t]
  \centering
  \includegraphics[width=1\columnwidth]{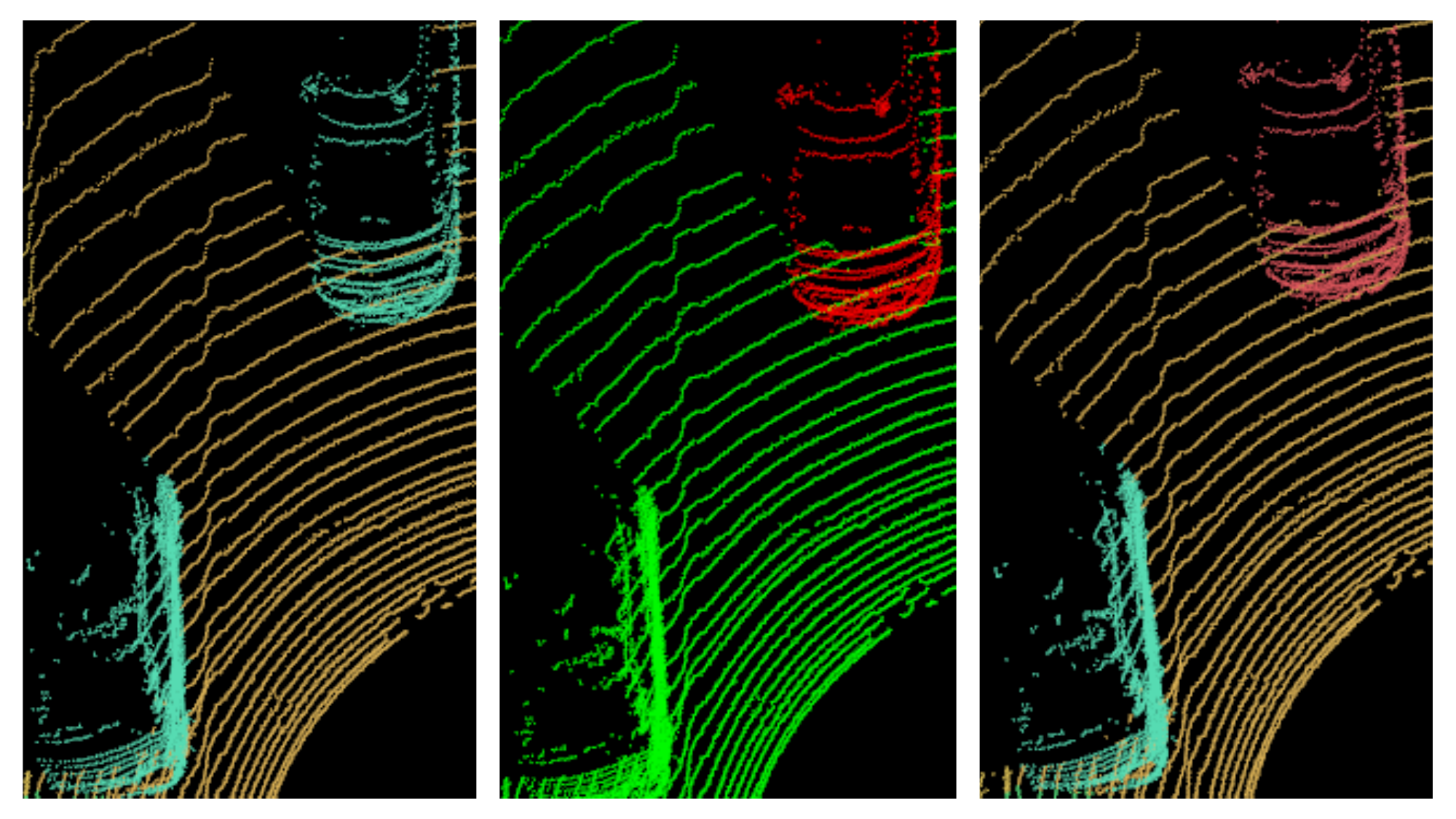}
  \caption{\textbf{Supervised Semantic Motion Labels.} The left semantic segmentation allows no distinction between the parked car at the bottom left and the moving car at the top right. The middle image shows a motion segmentation, where the parked car was classified as static, and the moving car as dynamic. Finally, the right image shows the fused semantic motion labels. Adapted from~\cite{Sartoris_Anomaly_2022_BA}.
  }
\label{fig:semantic_motion_labels}
\end{figure}

We derive semantic motion labels with a supervised \textbf{semantic segmentation} model~\cite{cortinhal_salsanext_2020} to determine whether a point belongs to a static or dynamic class. Some classes do not provide clear information about the motion state of the points, e.g., points assigned to the class \textit{cyclist} at a traffic light may be static in the case of a red light and dynamic in the case of a green light. By also performing supervised \textbf{motion segmentation}~\cite{chen_moving_2021}, we further subdivide classes into semantic motion labels, as shown in Fig.~\ref{fig:semantic_motion_labels}.


\subsection{Self-Supervised Predictive Motion Labels}
\label{subsec:self-supervised}

In order to predict motion labels for a given point cloud, we first filter out the ground~\cite{paigwar_gndnet_2020} to focus on objects in the scene, a common pre-processing step of \textbf{scene flow} models~\cite{wu2020pointpwc, mittal_just_2020, tishchenko_self-supervised_2020, kittenplon_flowstep3d_2021, baur_slim_2021}. Based on self-supervised flow prediction~\cite{kittenplon_flowstep3d_2021} of the remaining points, we aim to derive motion labels, indicating whether a point is static or dynamic. As shown in Fig.~\ref{fig:front_pic}, the model only performs predictions for points that are closer than 25 m and visible in the front RGB camera. The model takes consecutive point clouds as input and predicts the future motion for each lidar point in the form of a 3D displacement vector. The scene flow model does not distinguish between the point's own motion and the observer's ego-motion and represents the overall motion of a point between two consecutive frames. In order to derive relative displacements, we need to correct for the ego-motion. This can be done by leveraging or \textbf{learning odometry} information~\cite{nubert_self-supervised_2021}. After predicting the future point cloud \(\hat{X}_{t+1} = X_t + f_t
\), we apply the learned rigid body transformation \( T_{t+1 \rightarrow t}\) of an odometry model, transforming the predicted point cloud back into the coordinate system of \(X_{t}\). This gives the future point cloud \(\tilde{X}_{t+1}\), which contains only the predicted relative motion without the ego motion.

\begin{figure}[t]
  \centering
  \includegraphics[width=1.0\columnwidth]{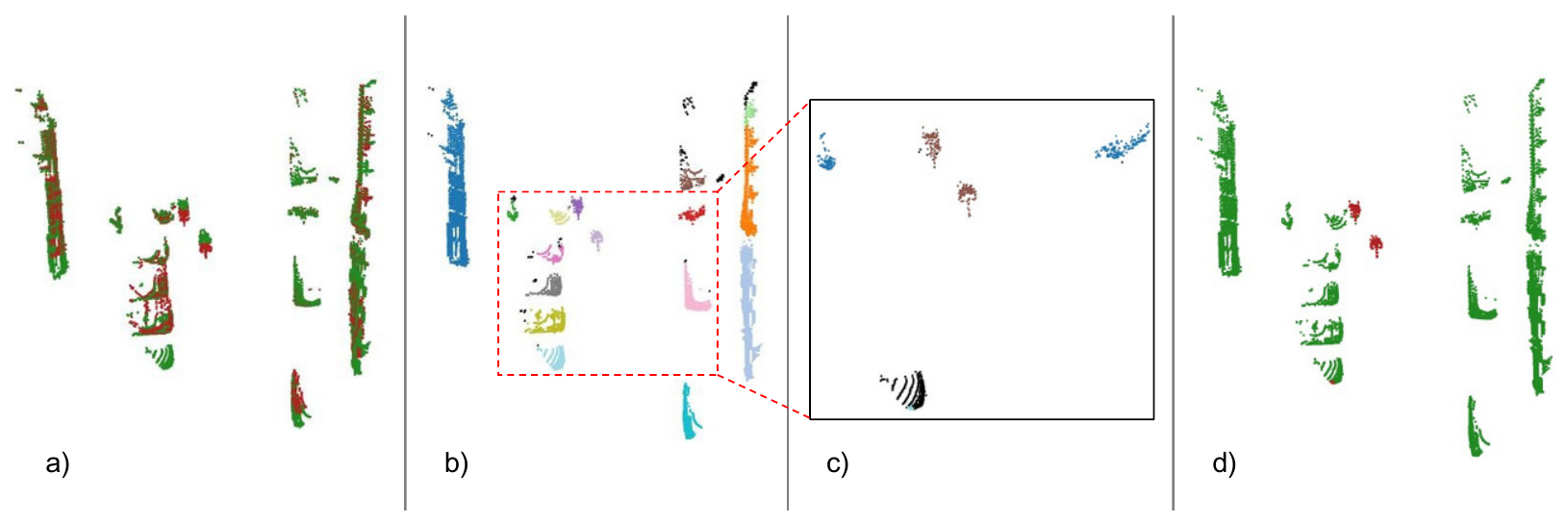}
  \caption{\textbf{Self-Supervised Predictive Motion Labels.} The first image shows the original point cloud in green and the point cloud transformed by the scene flow model and compensated by the ego-motion in red. The second and third images show the result of the spatial and flow-based clustering, respectively. The fourth image shows the final predictive motion labels, with dynamic points in red and static points in green. The scenes are shown from a Bird's-Eye View (BEV) perspective. Adapted from~\cite{Sartoris_Anomaly_2022_BA}.
  }
  \label{fig:motion_labels_ssv}
\end{figure}

As a result, static objects line up closely with the original data of \(X_t\), and only dynamic objects show a predicted displacement, as shown in Fig.~\ref{fig:motion_labels_ssv}a.

\begin{figure}[t!]
  \centering
  \includegraphics[width=1.0\columnwidth]{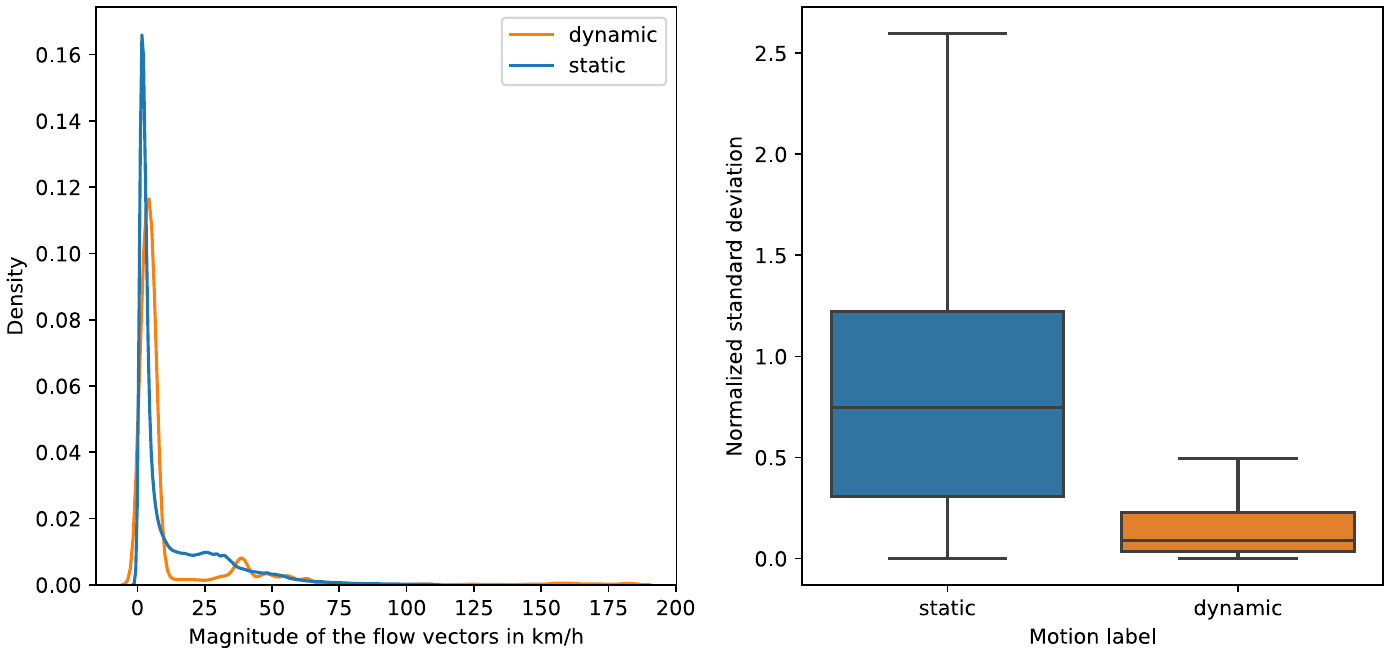}
  \caption{\textbf{Self-Supervised Label Generation}. The left graph shows that the magnitude of point-wise flow vectors is insufficient to distinguish between dynamic and static points. The boxplot on the right shows that the normalized standard deviation per instance is significantly lower for dynamic instances. Reprinted from~\cite{Sartoris_Anomaly_2022_BA}.
  }  \label{fig:boxplot_and_distribution}
\end{figure}

\textbf{Two-Stage Clustering.} An analysis of the velocity values of the flow predictions showed that separating static from dynamic classes is infeasible in a point-wise fashion, as a strong overlap exists. What we found, however, is a significant difference when considering instance-wise normalized standard deviations, as shown in Fig.~\ref{fig:boxplot_and_distribution}. While we performed this analysis with ground-truth labels, the necessity arises to form instance clusters during inference, as labels are not present in raw data.

As shown in Fig.~\ref{fig:motion_labels_ssv}b, we utilize DBSCAN~\cite{ester_density-based_1996} to cluster the point clouds spatially. A cluster is classified as potentially dynamic if the normalized standard deviation of the cluster's velocity is below 0.12, as identified through a grid search. To further reduce false positives, the potentially dynamic points are clustered based on their flow vectors. Points with a similar flow are clustered, causing fewer static clusters to be incorrectly classified as dynamic. This way, spatially separated points are grouped based on their flow, as shown in Fig.~\ref{fig:motion_labels_ssv}c. Here, the blue points on the left and right edges belonging to static objects now form a cluster, which changes the distribution of flow vectors within this cluster, causing it to be classified as static. Finally, the newly found clusters are classified as dynamic if the speed of the cluster is above 4 km/h, based on typical velocity profiles of pedestrians, as visible in Fig.~\ref{fig:motion_labels_ssv}d.

\subsection{Discrepancy Detection and Failure Classification}
\label{subsec:discrepancy}

After obtaining motion labels from both the \textit{supervised} and the \textit{self-supervised} stream, contradictions between the labels are detected, see Fig.~\ref{fig:method_overview}. Only the lidar points per frame for which both streams predicted a label are considered. Given a semantic and a predictive motion label for each lidar point, there exist four categories: Points which both models deem static~(green~{\color{green}$\bullet$}); points which both models deem dynamic~(blue~{\color{blue}$\bullet$}); points where the \textit{supervised} stream predicts a static point and the \textit{self-supervised} a dynamic one~(red~{\color{red}$\bullet$}), and points where the \textit{supervised} stream predicts a dynamic point and the \textit{self-supervised} a static one~(yellow~{\color{yellow}$\bullet$}). Finally, we cluster instances with contradicting labels so that an oracle, such as a human expert, can classify both complete scenes and single instances.

\subsection{Implementation Details}
\label{subsec:implementation}

For all models shown in Fig.~\ref{fig:method_overview}, we utilized publicly available models and model architectures to demonstrate the modularity of our approach. We trained the supervised semantic segmentation model SalsaNext~\cite{cortinhal_salsanext_2020} and the supervised motion segmentation model of Chen et al.~\cite{chen_moving_2021} on the KITTI-360 dataset~\cite{liao_kitti-360_2021,sanchez_recoverkitti360label_2022}, as it is a large dataset that contains semantic labels, motion labels, and odometry data. The training was performed on an NVIDIA RTX A6000. Hyperparameters were taken from the original papers~\cite{cortinhal_salsanext_2020, chen_moving_2021}. For the remaining three models, we used available pre-trained model weights. For ground segmentation, we deployed GndNet~\cite{paigwar_gndnet_2020}. For self-supervised scene flow estimation, we used FlowStep3D~\cite{kittenplon_flowstep3d_2021}. For the self-supervised odometry model, we deployed DeLORA~\cite{nubert_self-supervised_2021}. More details can be found in~\cite{Sartoris_Anomaly_2022_BA}.

\section{Evaluation}
\label{sec:eval}

Model failures can occur in seemingly normal situations~\cite{heideckerApplicationDrivenConceptualizationCorner2021,pfeilWhySystemMakes2022,zhou_data,heideckerCornerCaseDefinition}, but models are especially prone to failure in the presence of anomalies or outliers~\cite{breitensteinSystematizationCornerCases2020,breitenstein_corner_2021,heideckerApplicationDrivenConceptualizationCorner2021,bogdollDescriptionCornerCases2021,bogdollAnomalyDetectionAutonomous2022,vaterSystematicApproachDefinition2023,liuCurseRarityAutonomous2024}. In Section~\ref{subsec:qualitative}, we first analyze our approach given regular data. Here, no ground truth for especially challenging situations is available. Thus, we perform a qualitative evaluation by manually analyzing our method on over 20,000 frames.

In Section~\ref{subsec:quantitative}, we focus on anomalies. Here, ground truth is available. As CODA~\cite{li_coda_2022}, the largest real-world dataset with anomalies at the time of writing, only provides labels in RGB space, we first introduce \textit{LidarCODA}, an extension that provides labels in lidar space. Subsequently, we perform a quantitative evaluation to better understand the sensitivity of our method towards outliers.

\begin{figure}[t!]
  \centering
  \includegraphics[width=1.0\columnwidth]{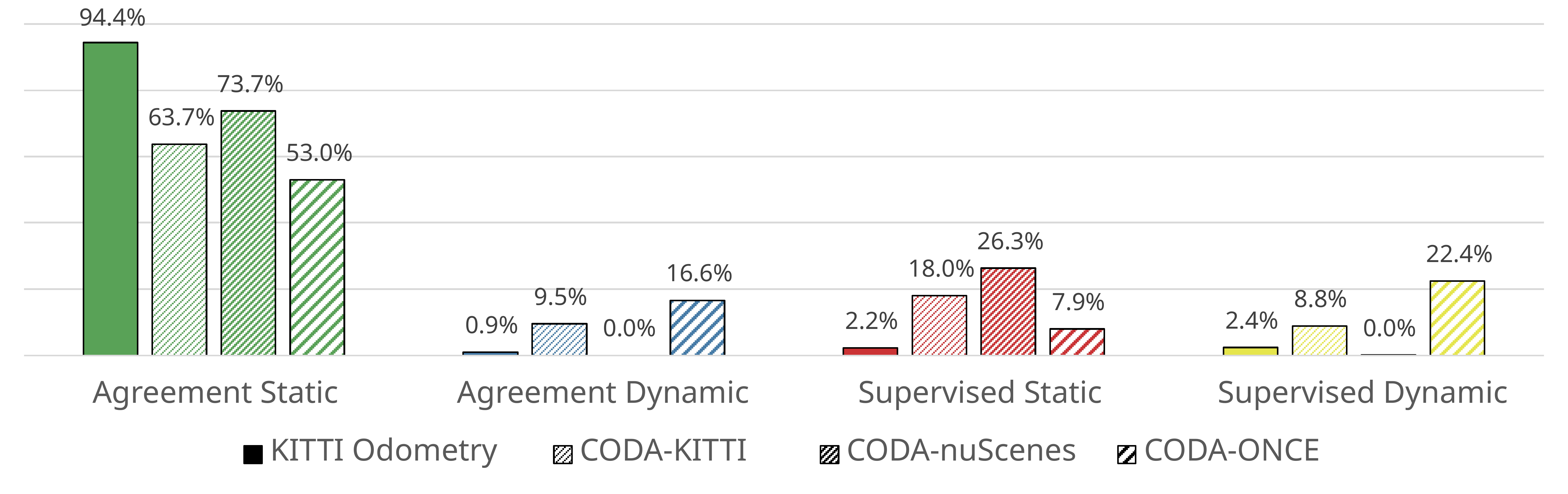}
   \caption{\textbf{Discrepancy Detection.} The charts show the distribution of the four possible outcomes of the discrepancy detection module for four different datasets. Given mostly normal data in the KITTI Odometry dataset, the two streams agree on 95~\% of the data (green,~blue). For outlier scenarios, as provided by LidarCODA, many more discrepancies (red,~yellow) are detected.}
   \label{fig:eval_stats}
\end{figure}

Fig.~\ref{fig:eval_stats} provides an overview of the results of discrepancy detection, the second-to-last step of our approach, for multiple datasets. For \textbf{regular scenarios}, the majority of points are predicted as static by both models and around 5 \% of the points show model contradictions. For \textbf{outlier scenarios}, many more disagreements take place. However, a large variety among the subsets of \textit{LidarCODA} can be observed. In the following, we first qualitatively examine scenarios in which the models disagree. Subsequently, we quantitatively examine the correlation between disagreements and anomalies in the environment.

\subsection{Regular Scenarios}
\label{subsec:qualitative}

\begin{figure}[t!]
  \centering
  \includegraphics[width=1.0\columnwidth]{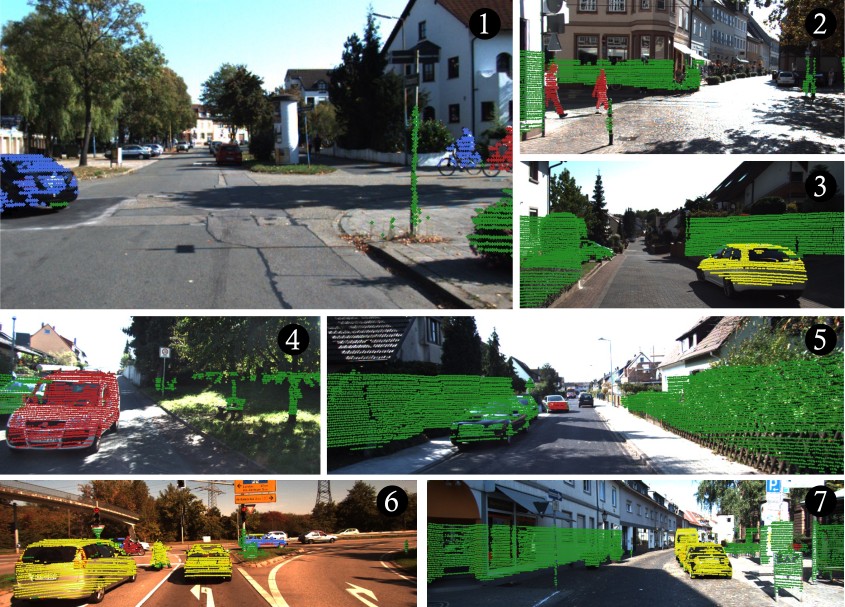}
   \caption{\textbf{Supervised Model Failures}. These exemplary images show model failures of the \textit{supervised} stream, which can be detected due to contradicting outputs of the \textit{self-supervised} model. Adapted from~\cite{Sartoris_Anomaly_2022_BA}.
   }
\label{fig:anomalies_detected_by_SSV}
\end{figure}

Regular scenarios represent the majority of miles driven during deployment, and it is important to understand situations in which models fail. However, the evaluation under regular scenarios is challenging, as no ground truth is available. Thus, we manually examined our approach on over 20,000 frames. This includes manually inspecting projected lidar point clouds on the front RGB image and assessing whether the classifications of our approach are contradictory, as explained in Sec.~\ref{subsec:discrepancy} and shown, for example, in Fig.~\ref{fig:anomalies_detected_by_SSV}. By human assessment, we determine if an object that was deemed static or dynamic by the models was actually static or dynamic, which is possible as multiple frames forming a temporal scenario are available for each frame. This way, we determine which model was wrong in cases of disagreement and can also spot cases where both models are wrong but agree.

\textbf{Evaluation Data.} For training, KITTI-360 and several sequences of the KITTI Odometry were used. To minimize perceptual failures due to a domain shift, we perform the qualitative evaluation on the remaining 20,350 frames of the KITTI Odometry sequences 11-21. The datasets are closely related, as both were captured in Karlsruhe, Germany with a Velodyne HDL-64E lidar.

\textbf{Evaluation.} As shown in Fig.~\ref{fig:method_overview}, the final stage of our approach is the classification of model failure modes. Our qualitative evaluation is performed by a human oracle. For visual inspection, we utilize lidar points mapped onto the corresponding RGB image for an improved scene understanding, as shown in Fig.~\ref{fig:anomalies_detected_by_SSV} with the color scheme introduced in Section~\ref{subsec:discrepancy}. In most cases, both streams were correctly consistent. In the following, we qualitatively present representative examples of detected model failures and highlight those that occurred frequently, suggesting general model flaws.

First, we discuss model failures of the supervised stream, which is the model under test in most cases. Here, the two streams contradict each other. We show representative examples in Fig.~\ref{fig:anomalies_detected_by_SSV}. Scene 1 shows a turning car and two moving bicyclists, where one bicyclist is wrongly labeled as static by the supervised stream. Scene 2 contains two walking pedestrians that are wrongly classified as static by the supervised stream. Scene 3 shows a parked car misclassified as dynamic by the supervised stream. Scenes 4 and 5 show a car moving slowly and a car moving backward, respectively. These cases demonstrate effectively that our approach enables the detection of regular but challenging scenarios that lead to model failures. Such model failures remain undetected in small evaluation datasets. We find various weak points in each stream, characterized by repeated occurrence. Specifically, the supervised model under test has weaknesses in distinguishing between dynamic and static objects in specific situations, e.g., at red lights or when a car is parked directly in front of the ego vehicle. Examples of such situations are given in scenes 6 and 7 of Fig.~\ref{fig:anomalies_detected_by_SSV}. 

\begin{figure}[t!]
  \centering
  \includegraphics[width=1.0\columnwidth]{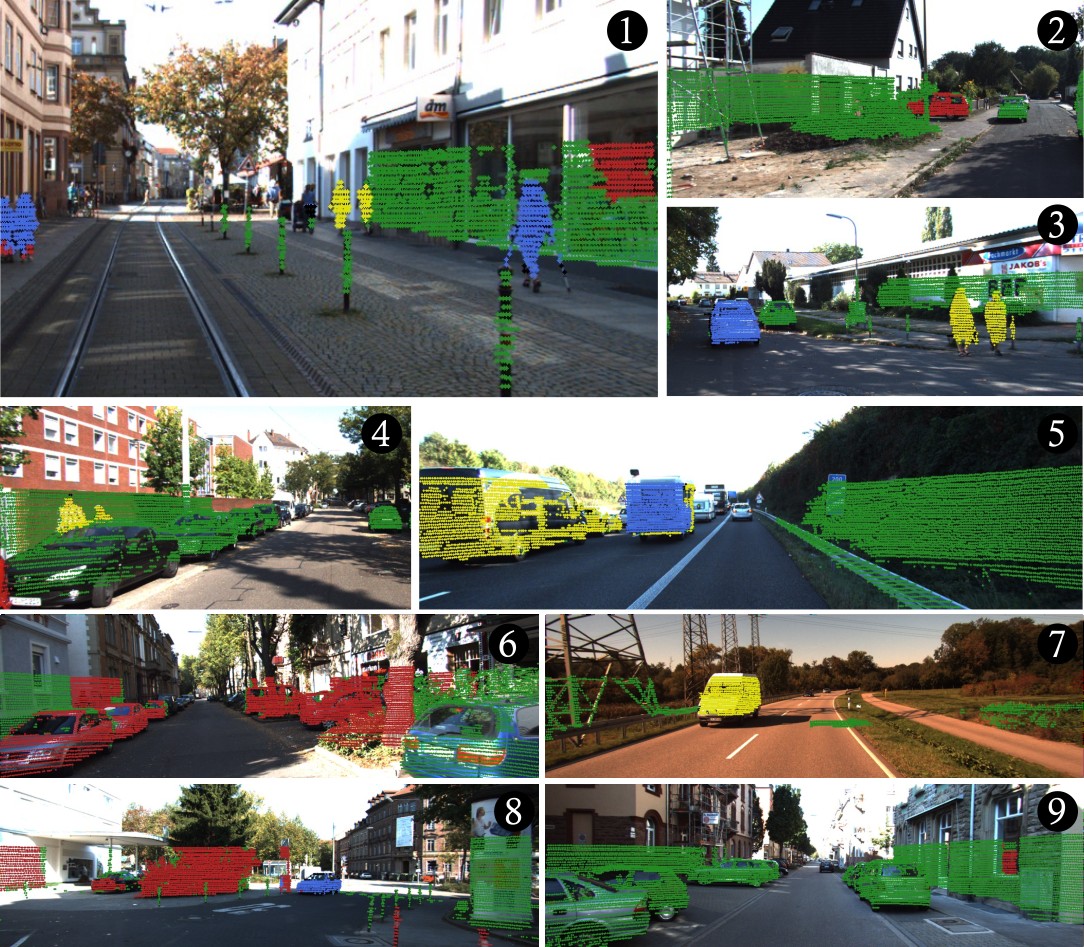}
   \caption{\textbf{Self-Supervised Model Failures}. These exemplary images show model failures of the \textit{self-supervised} model, which can be detected due to contradicting outputs of the \textit{supervised} stream. Adapted from~\cite{Sartoris_Anomaly_2022_BA}.
   }
\label{fig:anomalies_detected_by_SV}
\end{figure}

Next, we analyze scenarios where self-supervised model failures occur, detected by correct predictions of the supervised stream. Fig.~\ref{fig:anomalies_detected_by_SV} shows representative scenes. Scene 1 contains two distant pedestrians walking, wrongly classified as static by the self-supervised stream. In scene 2, a parked car is misclassified as static by the self-supervised stream. Scenes 3, 4, and 5 show walking pedestrians or moving cars incorrectly classified as dynamic. These cases demonstrate that our approach enables the detection of challenging temporal scenarios. The self-supervised stream classifies an above-average number of objects as dynamic when the ego-vehicle turns or goes over speed bumps. An example is shown in Fig.~\ref{fig:anomalies_detected_by_SV}, where in scene 6, the vehicle turns, and in scene 8, it drives over a speed bump. Another weak point is fast oncoming vehicles on highways, often classified as static, as seen in scene 7. Finally, a common weakness is small clusters on the side, which are incorrectly classified as dynamic, as in scene 9, where a window is classified as dynamic. 

\begin{figure}[b]
  \centering
  \includegraphics[width=1.0\columnwidth]{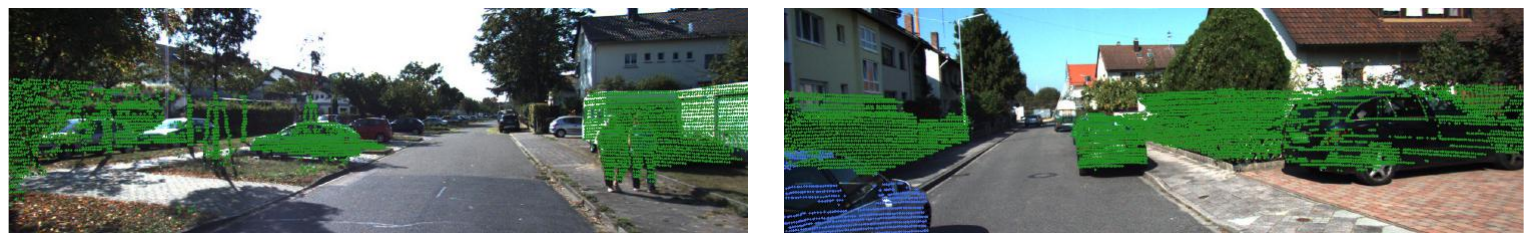}
   \caption{\textbf{Simultaneous Model Failures.} Examples where both streams produce model failures. Both cases are misclassified and are, therefore, consistent. Reprinted from~\cite{Sartoris_Anomaly_2022_BA}.
   }
   \label{fig:incorrectly_consistent}
\end{figure}

In rare cases, both models are incorrectly consistent, i.e., both streams agree, but the label is incorrect in both cases. We show examples in Fig.~\ref{fig:incorrectly_consistent}. Here, the left scene shows two walking pedestrians that are incorrectly classified as static, and the right scene shows a parked car that is classified as dynamic. 

\subsection{Outlier Scenarios}
\label{subsec:quantitative}

Scenarios with anomalies or outliers are known to lead to model failures~\cite{bogdoll_anomaly_2022}. Based on state-of-the-art datasets, evaluating lidar-based anomaly detection models has been challenging. Evaluation datasets are either unavailable~\cite{wong_identifying_2020,nunes_unsupervised_2022} or utilize known classes but exclude them from training data~\cite{cen_open-set_2021}. Thus, we first introduce \textit{LidarCODA}, a real-world anomaly dataset with labeled anomalies in lidar space. Subsequently, we quantitatively evaluate how sensitive our approach is to anomalies in the environment.

\textbf{Evaluation Data.} For the evaluation, we utilized data from the CODA dataset~\cite{li_coda_2022}, the only publicly available real-world dataset that provides lidar data and includes anomalies~\cite{bogdoll_perception}. The CODA dataset provides anomaly labels for objects based on three existing datasets: KITTI~\cite{geiger_are_2012}, ONCE~\cite{mao2021million}, and nuScenes~\cite{nuscenes}. CODA defines an anomaly as an object that \enquote{blocks or is about to block a potential path of the self-driving vehicle}~\cite{li_coda_2022} and/or \enquote{does not belong to any of the common classes of autonomous driving benchmarks}~\cite{li_coda_2022}. While the first risk-aware definition is not always in line with the methodology of our approach, where objects that block the path in front of the ego vehicle are not necessarily hard to segment or predict, the second one is well-suited. Novel classes are often more challenging for supervised methods compared to self-supervised approaches.

For the CODA-KITTI split, the authors of CODA manually reviewed all \textit{misc} labels available in the ground truth and relabeled some as anomalies according to a labeling policy. This enables us to examine our approach quantitatively with only a small domain gap. For CODA-nuScenes, the authors similarly adopted available annotations in a manual process. Finally, for CODA-ONCE, they deployed an automated anomaly detection approach, making this subset the most relevant. CODA includes 1,500 scenes with a total of 5,937 anomaly instances. Among those, 4,746 belong to the superclass \textit{traffic\_facility}, followed by 929 \textit{vehicle} and 197 \textit{obstruction} instances. With 396, most \textit{vehicle} instances can be found in CODA-KITTI.

\begin{figure}[b]
    \centering
    \begin{minipage}{\dimexpr\linewidth/3\relax}
        \centering
        \includegraphics[width=0.98\linewidth]{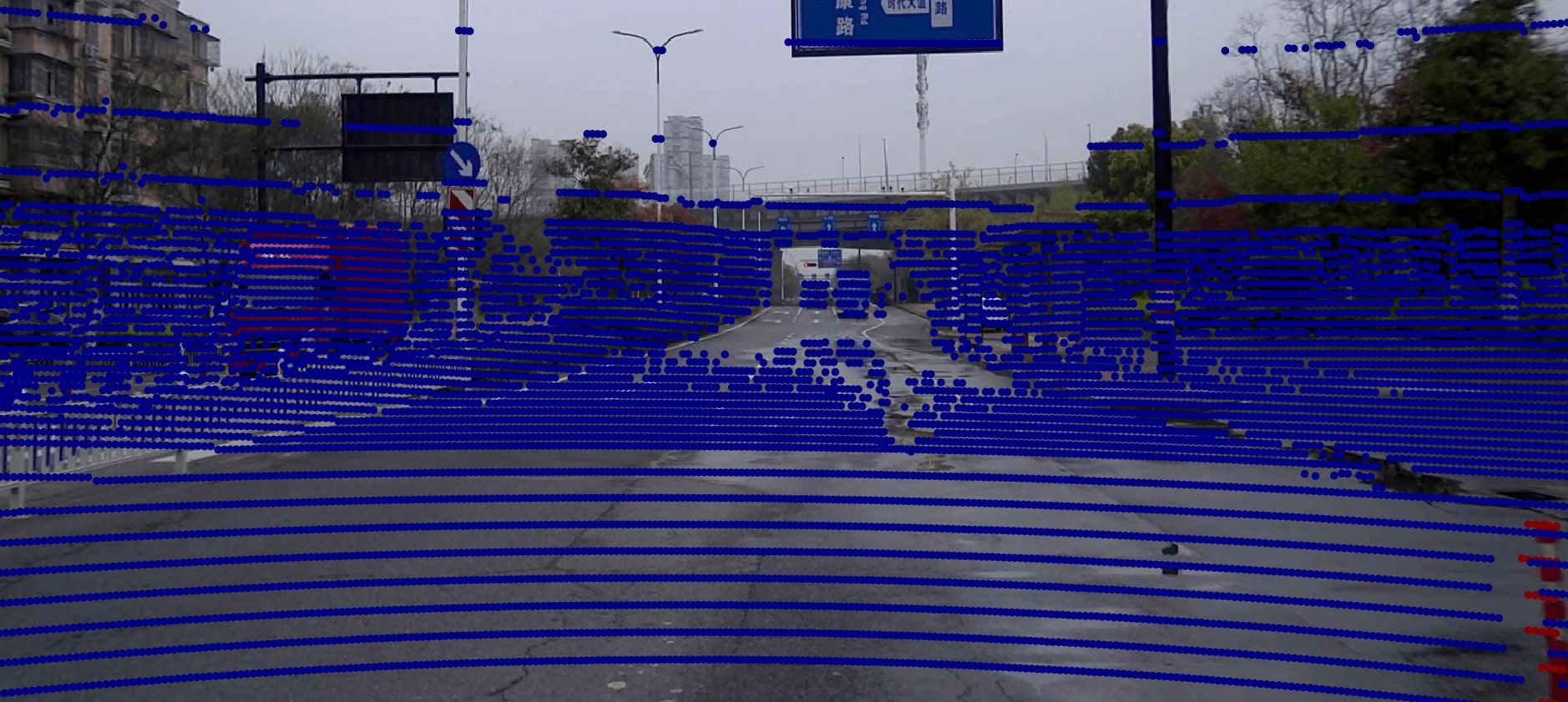}%
    \end{minipage}%
    \begin{minipage}{\dimexpr\linewidth/3\relax}
        \centering
        \includegraphics[width=0.98\linewidth]{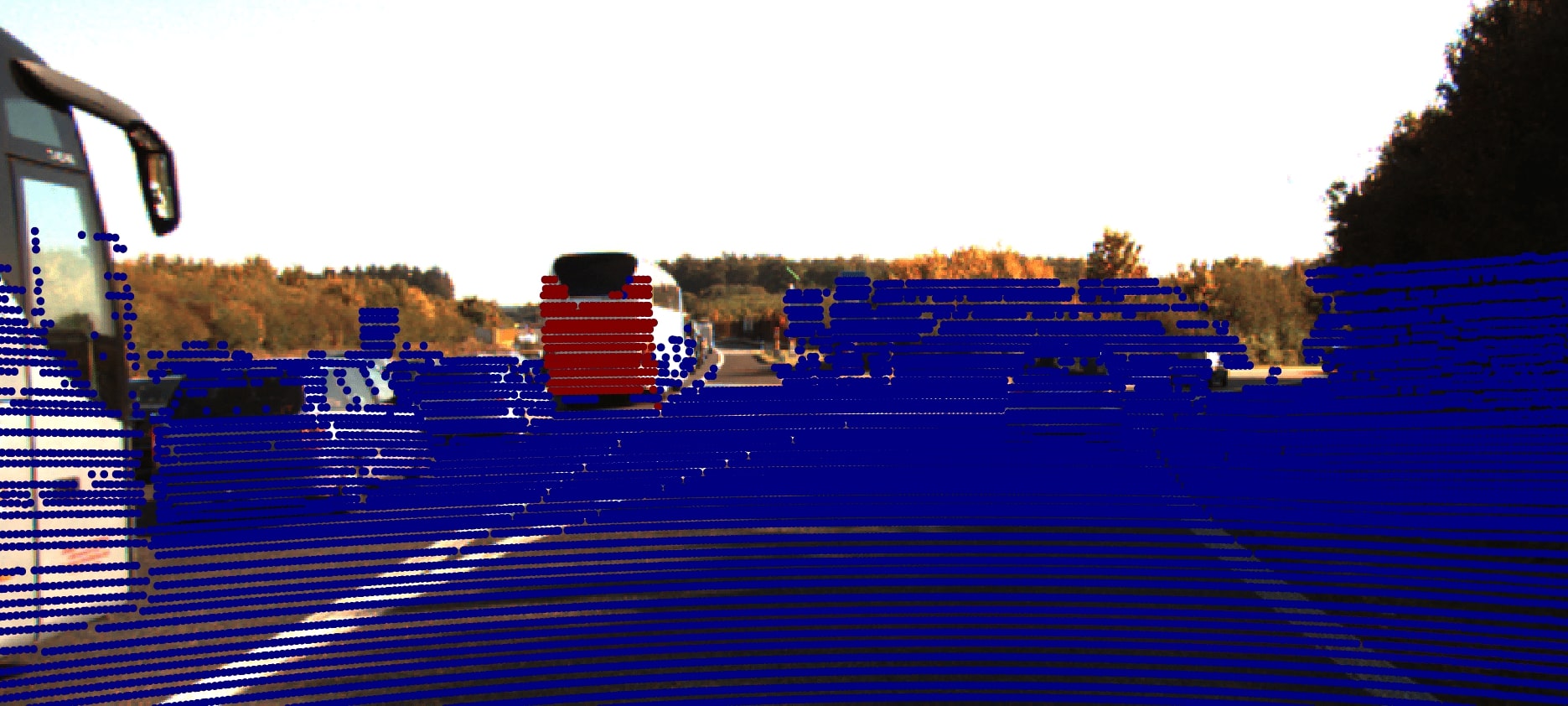}%
    \end{minipage}%
    \begin{minipage}{\dimexpr\linewidth/3\relax}
        \centering
        \includegraphics[width=0.98\linewidth]{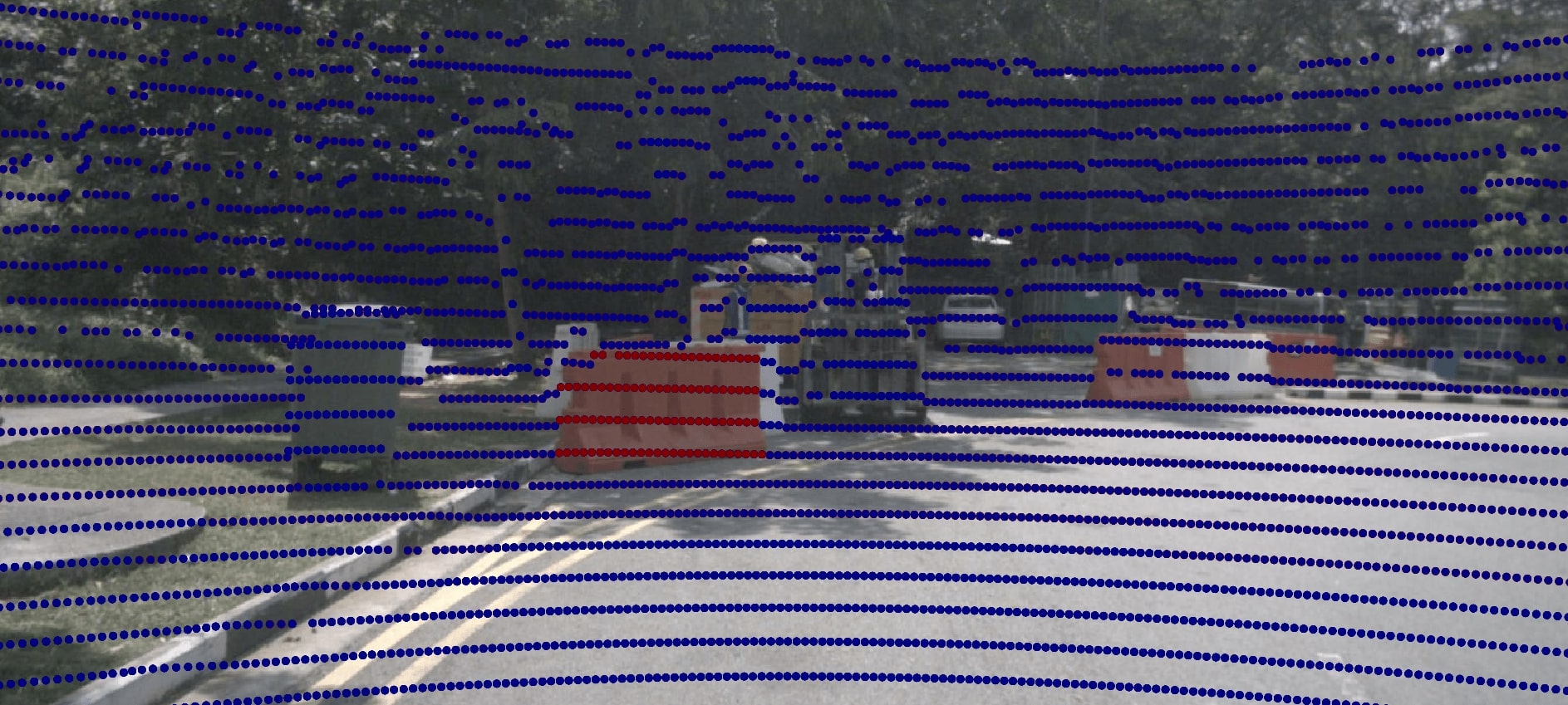}%
    \end{minipage}%
    \caption{\textbf{LidarCODA.} Annotated lidar scenes in the three data splits ONCE, KITTI, and nuScenes, from left to right. Anomalies are shown in red. Reprinted from~\cite{Geppert_Anomaly_2023_BA}.
    }
    \label{fig:lidarcoda}
\end{figure}

The CODA dataset provides anomaly labels only in the form of 2D bounding boxes in image space. However, point-wise labels in 3D lidar space are necessary to utilize CODA for our approach. Therefore, we present \textit{LidarCODA}, a dataset based on the CODA dataset~\cite{li_coda_2022} for evaluation. Based on a frustum-based filter, subsequent clustering, and manual inspection, we transfer the original 2D labels from image space into refined, point-wise 3D labels that go beyond the coarse characteristic of the provided bounding boxes. More details can be found in~\cite{Geppert_Anomaly_2023_BA}. \textit{LidarCODA} is the first real-world anomaly dataset for \textit{object-level} anomalies~\cite{breitensteinSystematizationCornerCases2020} with annotated lidar data, as shown in Fig.~\ref{fig:lidarcoda}. Here, the different lidar systems utilized also become clearly visible. Due to the sparse point cloud of nuScenes, many small or distant labeled anomalies in the image space are only covered by a few or no lidar points.

\begin{table}[t!]
\caption{\textbf{LidarCODA Subsets.} Evaluation of our approach on LidarCODA and its three subsets. Best results \textbf{bold} and second-best \underline{underlined}. Adapted from~\cite{Geppert_Anomaly_2023_BA}.
}
\resizebox{\columnwidth}{!}{%
\begin{tabular}{@{}llcccc@{}}
\toprule
\textbf{Dataset}   & \textbf{\#Frames} & \textbf{mIoU} $\uparrow$ & \textbf{AP} $\uparrow$   & \textbf{AR} $\uparrow$   & \textbf{F1} $\uparrow$   \\ \midrule
LidarCODA          & 1,412             & 8.9           & 13.2          & 26.2          & 17.5          \\ \midrule
LidarCODA-ONCE     & 1,034             & \underline{8.9}           & \textbf{14.0} & \underline{27.1}          & \textbf{18.4} \\
LidarCODA-KITTI    & 307               & \textbf{10.9} & \underline{13.3}          & \textbf{29.0} & \underline{18.3}          \\
LidarCODA-nuScenes & 71                & 0.4           & 0.7           & 1.0           & 0.8           \\ \bottomrule
\end{tabular}%
}

\label{tab:eval}
\end{table}

\textbf{Evaluation.} To quantitatively evaluate the correlation between model disagreements and anomalies present in the environment, we utilize the standard metrics mean Intersection over Union (mIoU), Average Precision (AP), Average Recall (AR), and F1 score, as shown in Tables~\ref{tab:eval} and~\ref{tab:eval_superclasses}. For a fair evaluation, we consider all points of the lidar point cloud, even if our approach did not label individual points, e.g., because they were filtered out during pre-processing. Such cases were counted as false negatives if an anomaly was missed.

To better understand the suitability of LidarCODA under introduced domain shifts, either due to new environments or due to new sensor setups, we perform experiments on the individual subsets, as shown in Table~\ref{tab:eval}. The results clearly show that our approach struggles with the nuScenes subset, which is primarily due to the large domain shift with respect to the sensor setup. Our approach is more sensitive towards anomalies for the subsets ONCE and KITTI. This is reflected in Fig.~\ref{fig:eval_stats}, where the CODA subsets also show much higher detection rates of model failures compared to the analysis with regular scenarios. This aligns with the much higher number of anomalies, even though the subsets reveal strongly varying behavior patterns.

Next, we investigate the sensitivity of our approach towards the superclasses provided in CODA by evaluating points that have labels assigned by both our approach and the ground truth. As shown in Table~\ref{tab:eval_superclasses}, our approach shows different levels of sensitivity given different types of anomalies, being most sensitive to \textit{cyclists} and objects of the \textit{misc} class. The model performs worst on the class \textit{animal}, which is difficult to interpret given the small number of only five instances. The \textit{misc} class contains objects that are \enquote{unrecognizable or difficult to categorize}~\cite{li_coda_2022}. These results align well with our approach, where \textit{cyclist} instances, which are hard to predict by the self-supervised stream, and \textit{misc} instances, which are rare and thus hard to classify by the supervised stream, lead to model disagreements, as the complementary model does not display similar weaknesses in both cases.

\section{Conclusion}
\label{sec:concl}

\begin{table}[t]
\centering
\caption{\textbf{CODA Superclasses}. Evaluation of our approach on seven superclasses. Best results \textbf{bold} and second-best \underline{underlined}. Adapted from~\cite{Geppert_Anomaly_2023_BA}.
}
\resizebox{\columnwidth}{!}{%
\begin{tabular}{@{}llcccc@{}}
\toprule
\textbf{Superclass} & \textbf{\#Instances} & \textbf{mIoU} $\uparrow$ & \textbf{AP} $\uparrow$  & \textbf{AR} $\uparrow$  & \textbf{F1} $\uparrow$  \\ \midrule
Pedestrian          & 16                   & 33.9          & 44.1          & 37.3          & 40.4          \\
Cyclist             & 22                   & \textbf{41.6} & \underline{58.3}          & \textbf{49.5} & \textbf{53.5} \\
Vehicle             & 736                  & 33.0          & 48.3          & \underline{41.0}          & 44.4          \\
Animal              & 5                    & 0.0           & 0.0           & 0.0           & 0.0           \\
Traffic facility    & 3,360                 & 28.6          & 39.9          & 33.9          & 36.7          \\
Obstruction         & 125                  & 20.5          & 34.0          & 22.9          & 27.4          \\
Misc                & 15                   & \underline{36.7}          & \textbf{60.7} & 37.9          & \underline{46.7}          \\ \bottomrule
\end{tabular}%
}

\label{tab:eval_superclasses}
\end{table}

In this work, we have presented a label-free approach to detect model failures for the segmentation of point clouds, focusing on the classes \textit{static} and \textit{dynamic}. We leverage complementary learning paradigms to detect contradicting outputs on the same task, consisting of a \textit{supervised} stream for semantic motion labels and a \textit{self-supervised} stream for predictive motion labels. This way, we detect model failures far beyond the limited scope of a small evaluation dataset. We first inspect model failures in regular scenarios. By manually analyzing over 20,000 frames qualitatively, we detect model failures in seemingly normal scenarios and are able to categorize frequently occurring cases. In regular scenarios, our method categorizes 95 \% of the data as typical, which makes human analysis of the remaining 5 \% feasible even for larger datasets.

Second, we analyzed the sensitivity of our approach towards scenarios with anomalies, as defined by the CODA dataset, and showed a moderate sensitivity. In order to quantitatively examine our approach, we introduced \textit{LidarCODA}, the first real-world dataset with labeled anomalies in lidar space. We demonstrated that our approach shows an increased sensitivity to often hard-to-detect and hard-to-predict bicycles as well as hard-to-categorize objects.

Our approach effectively unveils model failures far beyond those that can be detected with small evaluation datasets. This leads to an increased understanding of the model performance in large-scale deployments, leveraging abundantly available unlabeled data. Model failures detected by our approach can be utilized to collect additional training data representing both static and temporally challenging scenarios.

\textbf{Limitations.} When both streams are wrong, model failures go undetected. This behavior is known and unavoidable~\cite{wangLearningDisagreementEvent2022,wangLearningFromDisagreementModelComparison2023} and can be mitigated by deploying multiple approaches or triggers to detect challenging scenarios~\cite{wadatcvprCVPR21WAD2021}. In addition, our self-supervised stream depends on a clustering strategy, which can lead to faulty clusters and misclassification.

\textbf{Future Work.} The proposed \textit{complementary learning} approach can be applied to the detection of model failures in different settings that can be addressed with different training paradigms, such as full panoptic segmentation beyond the utilized static and dynamic classes~\cite{niu2024unsupervised}, lane detection~\cite{nie2024lanecorrectselfsupervisedlanedetection}, or drivable-area segmentation~\cite{mayr_ssl_drivable}.

\section*{Acknowledgment}
\label{sec:ackno}
This work results partly from the just better DATA project supported by the German Federal Ministry for Economic Affairs and Climate Action (BMWK), grant number 19A23003H.

\bibliographystyle{IEEEtran}
\bibliography{references}

\begin{thebibliography}{10}
\providecommand{\url}[1]{#1}
\csname url@samestyle\endcsname
\providecommand{\newblock}{\relax}
\providecommand{\bibinfo}[2]{#2}
\providecommand{\BIBentrySTDinterwordspacing}{\spaceskip=0pt\relax}
\providecommand{\BIBentryALTinterwordstretchfactor}{4}
\providecommand{\BIBentryALTinterwordspacing}{\spaceskip=\fontdimen2\font plus
\BIBentryALTinterwordstretchfactor\fontdimen3\font minus \fontdimen4\font\relax}
\providecommand{\BIBforeignlanguage}[2]{{%
\expandafter\ifx\csname l@#1\endcsname\relax
\typeout{** WARNING: IEEEtran.bst: No hyphenation pattern has been}%
\typeout{** loaded for the language `#1'. Using the pattern for}%
\typeout{** the default language instead.}%
\else
\language=\csname l@#1\endcsname
\fi
#2}}
\providecommand{\BIBdecl}{\relax}
\BIBdecl

\bibitem{nuscenes}
H.~Caesar, V.~Bankiti, A.~H. Lang, S.~Vora \emph{et~al.}, ``nuscenes: A multimodal dataset for autonomous driving,'' in \emph{Conference on Computer Vision and Pattern Recognition (CVPR)}, 2020.

\bibitem{Sun_2020_CVPR}
P.~Sun, H.~Kretzschmar, X.~Dotiwalla, A.~Chouard \emph{et~al.}, ``Scalability in perception for autonomous driving: Waymo open dataset,'' in \emph{Conference on Computer Vision and Pattern Recognition (CVPR)}, 2020.

\bibitem{Yu2018BDD100KAD}
F.~Yu, H.~Chen, X.~Wang, W.~Xian \emph{et~al.}, ``Bdd100k: A diverse driving dataset for heterogeneous multitask learning,'' in \emph{Conference on Computer Vision and Pattern Recognition (CVPR)}, 2020.

\bibitem{forbes_60_million}
R.~Bishop, ``60 million miles and counting: Robotaxis shift into high gear,'' \url{https://www.forbes.com/sites/richardbishop1/2024/07/27/60-million-miles-and-counting-robotaxis-shift-into-high-gear/}, 2024, accessed: 2025-01-29.

\bibitem{mao2021million}
J.~Mao, M.~Niu, C.~Jiang, H.~Liang \emph{et~al.}, ``One million scenes for autonomous driving: Once dataset,'' in \emph{Conference on Neural Information Processing Systems (NeurIPS)}, 2021.

\bibitem{Han2021SODA10MS}
J.~Han, X.~Liang, H.~Xu, K.~Chen \emph{et~al.}, ``Soda10m: A large-scale 2d self/semi-supervised object detection dataset for autonomous driving,'' in \emph{Conference on Neural Information Processing Systems (NeurIPS)}, 2021.

\bibitem{li2024multiagentmultitraversalmultimodalselfdriving}
Y.~Li, Z.~Li, N.~Chen, M.~Gong \emph{et~al.}, ``Multiagent multitraversal multimodal self-driving: Open mars dataset,'' in \emph{Conference on Computer Vision and Pattern Recognition (CVPR)}, 2024.

\bibitem{kumarActiveLearningQuery2020}
P.~Kumar and A.~Gupta, ``Active {{Learning Query Strategies}} for {{Classification}}, {{Regression}}, and {{Clustering}}: {{A Survey}},'' \emph{Journal of Computer Science and Technology}, vol.~35, no.~4, 2020.

\bibitem{wadatcvprCVPR21WAD2021}
A.~Karpathy, ``Tesla keynote: Cvpr 2021 workshop on autonomous driving,'' \url{https://www.youtube.com/watch?v=g6bOwQdCJrc}, 2021, accessed: 2024-06-13.

\bibitem{dengAreLabelsAlways2021}
W.~Deng and L.~Zheng, ``Are {{Labels Always Necessary}} for {{Classifier Accuracy Evaluation}}?'' in \emph{Conference on Computer Vision and Pattern Recognition (CVPR)}, 2021.

\bibitem{schubertIdentifyingLabelErrors2024}
M.~Schubert, T.~Riedlinger, K.~Kahl, D.~Kr{\"o}ll \emph{et~al.}, ``Identifying {{Label Errors}} in {{Object Detection Datasets}} by {{Loss Inspection}},'' in \emph{Winter Conference on Applications of Computer Vision (WACV)}, 2024.

\bibitem{cohnImprovingGeneralizationActive1994}
D.~Cohn, L.~Atlas, and R.~Ladner, ``Improving generalization with active learning,'' \emph{Machine Learning}, vol.~15, no.~2, 1994.

\bibitem{dietterich2000ensemble}
T.~G. Dietterich, ``Ensemble methods in machine learning,'' in \emph{International workshop on multiple classifier systems}.\hskip 1em plus 0.5em minus 0.4em\relax Springer, 2000.

\bibitem{pang2019improving}
T.~Pang, K.~Xu, C.~Du, N.~Chen, and J.~Zhu, ``Improving adversarial robustness via promoting ensemble diversity,'' in \emph{International Conference on Machine Learning (ICML)}, 2019.

\bibitem{lakshminarayanan2017simple}
B.~Lakshminarayanan, A.~Pritzel, and C.~Blundell, ``Simple and scalable predictive uncertainty estimation using deep ensembles,'' in \emph{Conference on Neural Information Processing Systems (NeurIPS)}, 2017.

\bibitem{mehrtens2022improving}
H.~A. Mehrtens, C.~Gonz{\'{a}}lez, and A.~Mukhopadhyay, ``Improving robustness and calibration in ensembles with diversity regularization,'' in \emph{German Conference on Pattern Recognition (GCPR)}, 2022.

\bibitem{pagliardiniAGREEDISAGREEDIVERSITY2023}
M.~Pagliardini, M.~Jaggi, F.~Fleuret, and S.~P. Karimireddy, ``Agree to disagree: Diversity through disagreement for better transferability,'' in \emph{International Conference on Learning Representations (ICLR)}, 2023.

\bibitem{wangLearningDisagreementEvent2022}
L.~Wang, J.~Wang, Y.~Zheng, S.~Jain \emph{et~al.}, ``Learning from {{Disagreement}} for {{Event Detection}},'' in \emph{{{International Conference}} on {{Big Data}}}, 2022.

\bibitem{kuncheva2003measures}
L.~I. Kuncheva and C.~J. Whitaker, ``Measures of diversity in classifier ensembles and their relationship with the ensemble accuracy,'' \emph{Machine learning}, vol.~51, 2003.

\bibitem{jacobs1991adaptive}
R.~A. Jacobs, M.~I. Jordan, S.~J. Nowlan, and G.~E. Hinton, ``Adaptive mixtures of local experts,'' \emph{Neural computation}, vol.~3, no.~1, 1991.

\bibitem{pavlitskayaUsingMixtureExpert2020a}
S.~Pavlitskaya, C.~Hubschneider, M.~Weber, R.~Moritz \emph{et~al.}, ``Using {{Mixture}} of {{Expert Models}} to {{Gain Insights}} into {{Semantic Segmentation}},'' in \emph{Conference on Computer Vision and Pattern Recognition (CVPR) Workshops}, 2020.

\bibitem{pavlitskaya2022evaluating}
S.~Pavlitskaya, C.~Hubschneider, and M.~Weber, ``Evaluating mixture-of-experts architectures for network aggregation,'' in \emph{Deep Neural Networks and Data for Automated Driving: Robustness, Uncertainty Quantification, and Insights Towards Safety}, 2022.

\bibitem{schroderRevisitingUncertaintybasedQuery2022}
C.~Schr{\"o}der, A.~Niekler, and M.~Potthast, ``Revisiting {{Uncertainty-based Query Strategies}} for {{Active Learning}} with {{Transformers}},'' in \emph{Findings of the {{Association}} for {{Computational Linguistics}}}, 2022.

\bibitem{roitberg2018informed}
A.~Roitberg, Z.~Al{-}Halah, and R.~Stiefelhagen, ``Informed democracy: Voting-based novelty detection for action recognition,'' in \emph{British Machine Vision Conference (BMVC)}, 2018.

\bibitem{yahaya2019consensus}
S.~W. Yahaya, A.~Lotfi, and M.~Mahmud, ``A consensus novelty detection ensemble approach for anomaly detection in activities of daily living,'' \emph{Applied Soft Computing}, vol.~83, 2019.

\bibitem{hinoActiveLearningQuery2022}
H.~Hino and S.~Eguchi, ``Active {{Learning}} by {{Query}} by {{Committee}} with {{Robust Divergences}},'' \emph{Information Geometry}, vol.~6, 2023.

\bibitem{chenLearningEfficientObject2017a}
G.~Chen, W.~Choi, X.~Yu, T.~Han, and M.~Chandraker, ``Learning {{Efficient Object Detection Models}} with {{Knowledge Distillation}},'' in \emph{Conference on Neural Information Processing Systems (NeurIPS)}, 2017.

\bibitem{pengEnergybasedAutomatedModel2024}
R.~Peng, H.~Zou, H.~Wang, Y.~Zeng, Z.~Huang, and J.~Zhao, ``Energy-based {{Automated Model Evaluation}},'' in \emph{International Conference on Learning Representations (ICLR)}, 2024.

\bibitem{baekAgreementontheLinePredictingPerformance}
C.~Baek, Y.~Jiang, A.~Raghunathan, and Z.~Kolter, ``Agreement-on-the-{{Line}}: {{Predicting}} the {{Performance}} of {{Neural Networks}} under {{Distribution Shift}},'' in \emph{Conference on Neural Information Processing Systems (NeurIPS)}, 2022.

\bibitem{chenDetectingErrorsEstimating2021}
J.~Chen, F.~Liu, B.~Avci, X.~Wu, Y.~Liang, and S.~Jha, ``Detecting {{Errors}} and {{Estimating Accuracy}} on {{Unlabeled Data}} with {{Self-training Ensembles}},'' in \emph{Conference on Neural Information Processing Systems (NeurIPS)}, 2021.

\bibitem{wangLearningFromDisagreementModelComparison2023}
J.~Wang, L.~Wang, Y.~Zheng, C.-C.~M. Yeh, S.~Jain, and W.~Zhang, ``Learning-{{From-Disagreement}}: {{A Model Comparison}} and {{Visual Analytics Framework}},'' \emph{Transactions on Visualization and Computer Graphics}, vol.~29, no.~9, 2023.

\bibitem{yuPredictingOutofDistributionError2022}
Y.~Yu, Z.~Yang, A.~Wei, Y.~Ma, and J.~Steinhardt, ``Predicting {{Out-of-Distribution Error}} with the {{Projection Norm}},'' in \emph{International Conference on Machine Learning (ICML)}, 2022.

\bibitem{chachulaCombatingNoisyLabels2023}
K.~Chachula, J.~Lyskawa, B.~Olber, P.~Fratczak, A.~Popowicz, and K.~Radlak, ``Combating noisy labels in object detection datasets,'' \emph{arXiv:2211.13993}, 2023.

\bibitem{huProbabilityDifferentialBasedClass2022}
Z.~Hu, K.~Gao, X.~Zhang, J.~Wang, H.~Wang, and J.~Han, ``Probability {{Differential-Based Class Label Noise Purification}} for {{Object Detection}} in {{Aerial Images}},'' \emph{Geoscience and Remote Sensing Letters}, vol.~19, 2022.

\bibitem{barNovelBenchmarkRefinement2023}
A.~B{\"a}r, J.~Uhrig, J.~P. Umesh, M.~Cordts, and T.~Fingscheidt, ``A {{Novel Benchmark}} for {{Refinement}} of {{Noisy Localization Labels}} in {{Autolabeled Datasets}} for {{Object Detection}},'' in \emph{Conference on Computer Vision and Pattern Recognition (CVPR) Workshops}, 2023.

\bibitem{northcuttConfidentLearningEstimating2021}
C.~Northcutt, L.~Jiang, and I.~Chuang, ``Confident {{Learning}}: {{Estimating Uncertainty}} in {{Dataset Labels}},'' \emph{Journal of Artificial Intelligence Research}, vol.~70, 2021.

\bibitem{kowolYOdarUncertaintybasedSensor2020}
K.~Kowol, M.~Rottmann, S.~Bracke, and H.~Gottschalk, ``{{YOdar}}: {{Uncertainty-based Sensor Fusion}} for {{Vehicle Detection}} with {{Camera}} and {{Radar Sensors}},'' in \emph{International Conference on Agents and Artificial Intelligence (ICAART)}, 2021.

\bibitem{intelnewsroomCES2021Hood2021}
A.~Shashua, ``Intel newsroom: {{CES}} 2021: {{Under}} the {{Hood}},'' \url{https://youtube.com/watch?v=B7YNj66GxRA}, 2021, accessed: 2024-06-13.

\bibitem{Baur2024ECCV}
S.~Baur, F.~Moosmann, and A.~Geiger, ``Liso: Lidar-only self-supervised 3d object detection,'' in \emph{European Conference on Computer Vision (ECCV)}, 2024.

\bibitem{lentsch2024union}
T.~Lentsch, H.~Caesar, and D.~M. Gavrila, ``{UNION}: Unsupervised {3D} object detection using object appearance-based pseudo-classes,'' in \emph{Conference on Neural Information Processing Systems (NeurIPS)}, 2024.

\bibitem{shaoWhenItLikely2024}
W.~Shao, B.~Li, W.~Yu, J.~Xu, and H.~Wang, ``When {{Is It Likely}} to {{Fail}}? {{Performance Monitor}} for {{Black-Box Trajectory Prediction Model}},'' \emph{Transactions on Automation Science and Engineering}, 2024.

\bibitem{buerkleSafePerceptionHierarchical2022}
C.~Buerkle, F.~Oboril, J.~Burr, and K.-U. Scholl, ``Safe perception - a hierarchical monitor approach,'' in \emph{International Conference on Intelligent Transportation Systems (ITSC)}, 2022.

\bibitem{delicOutlierDetectionEnsembling2024c}
A.~Deli{\'c}, M.~Grci{\'c}, and S.~{\v S}egvi{\'c}, ``Outlier detection by ensembling uncertainty with negative objectness,'' in \emph{British Machine Vision Conference (BMVC)}, 2024.

\bibitem{chenAdversarialRobustnessSelfSupervised2020}
T.~Chen, S.~Liu, S.~Chang, Y.~Cheng, L.~Amini, and Z.~Wang, ``Adversarial {{Robustness}}: {{From Self-Supervised Pre-Training}} to {{Fine-Tuning}},'' in \emph{Conference on Computer Vision and Pattern Recognition (CVPR)}, 2020.

\bibitem{chengSelfsupervisedSupervisedJoint2021}
Y.~Cheng, W.~Wang, L.~Jiang, and W.~Macherey, ``Self-supervised and {{Supervised Joint Training}} for {{Resource-rich Machine Translation}},'' in \emph{International Conference on Machine Learning (ICML)}, 2021.

\bibitem{cortinhal_salsanext_2020}
T.~Cortinhal, G.~Tzelepis, and E.~E. Aksoy, ``Salsanext: Fast, uncertainty-aware semantic segmentation of lidar point clouds for autonomous driving,'' in \emph{Advances in Visual Computing}, 2020.

\bibitem{chen_moving_2021}
X.~Chen, S.~Li, B.~Mersch, L.~Wiesmann \emph{et~al.}, ``Moving {Object} {Segmentation} in {3D} {LiDAR} {Data}: {A} {Learning}-based {Approach} {Exploiting} {Sequential} {Data},'' \emph{Robotics and Automation Letters}, 2021.

\bibitem{Sartoris_Anomaly_2022_BA}
F.~Sartoris, ``{Anomaly Detection in Lidar Data by Combining Supervised and Self-Supervised Methods},'' Bachelor's Thesis, {Karlsruhe Institute of Technology (KIT)}, 2022.

\bibitem{paigwar_gndnet_2020}
A.~Paigwar, O.~Erkent, D.~Sierra-Gonzalez, and C.~Laugier, ``{GndNet}: {Fast} {Ground} {Plane} {Estimation} and {Point} {Cloud} {Segmentation} for {Autonomous} {Vehicles},'' in \emph{International Conference on Intelligent Robots and Systems (IROS)}, 2020.

\bibitem{wu2020pointpwc}
W.~Wu, Z.~Y. Wang, Z.~Li, W.~Liu, and L.~Fuxin, ``Pointpwc-net: Cost volume on point clouds for (self-) supervised scene flow estimation,'' in \emph{European Conference on Computer Vision (ECCV)}, 2020.

\bibitem{mittal_just_2020}
H.~Mittal, B.~Okorn, and D.~Held, ``Just go with the flow: Self-supervised scene flow estimation,'' in \emph{Conference on Computer Vision and Pattern Recognition (CVPR)}, 2020.

\bibitem{tishchenko_self-supervised_2020}
I.~Tishchenko, S.~Lombardi, M.~R. Oswald, and M.~Pollefeys, ``Self-supervised learning of non-rigid residual flow and ego-motion,'' in \emph{International Conference on 3D Vision (3DV)}, 2020.

\bibitem{kittenplon_flowstep3d_2021}
Y.~Kittenplon, Y.~C. Eldar, and D.~Raviv, ``Flowstep3d: Model unrolling for self-supervised scene flow estimation,'' in \emph{Conference on Computer Vision and Pattern Recognition (CVPR)}, 2021.

\bibitem{baur_slim_2021}
S.~Baur, D.~Emmerichs, F.~Moosmann, P.~Pinggera, B.~Ommer, and A.~Geiger, ``{SLIM}: {Self}-{Supervised} {LiDAR} {Scene} {Flow} and {Motion} {Segmentation},'' in \emph{International Conference on Computer Vision (ICCV)}, 2021.

\bibitem{nubert_self-supervised_2021}
J.~Nubert, S.~Khattak, and M.~Hutter, ``Self-supervised learning of lidar odometry for robotic applications,'' in \emph{International Conference on Robotics and Automation (ICRA)}, 2021.

\bibitem{ester_density-based_1996}
M.~Ester, H.-P. Kriegel, J.~Sander, and X.~Xu, ``A {Density}-{Based} {Algorithm} for {Discovering} {Clusters} in {Large} {Spatial} {Databases} with {Noise},'' in \emph{{International} {Conference} on {Knowledge} {Discovery} and {Data} {Mining}}, 1996.

\bibitem{liao_kitti-360_2021}
Y.~Liao, J.~Xie, and A.~Geiger, ``{KITTI}-360: A novel dataset and benchmarks for urban scene understanding in 2d and 3d,'' \emph{Pattern Analysis and Machine Intelligence (PAMI)}, 2022.

\bibitem{sanchez_recoverkitti360label_2022}
J.~Sanchez, ``recoverkitti360label,'' \url{https://github.com/JulesSanchez/recoverKITTI360label/pull/3}, 2022, accessed: 2024-06-26.

\bibitem{heideckerApplicationDrivenConceptualizationCorner2021}
F.~Heidecker, J.~Breitenstein, R.~Kevin, L.~Jonas \emph{et~al.}, ``An {{Application-Driven Conceptualization}} of {{Corner Cases}} for {{Perception}} in {{Highly Automated Driving}},'' in \emph{Intelligent Vehicles Symposium (IV)}, 2021.

\bibitem{pfeilWhySystemMakes2022}
J.~Pfeil, J.~Wieland, T.~Michalke, and A.~Theissler, ``On {{Why}} the {{System Makes}} the {{Corner Case}}: {{AI-based Holistic Anomaly Detection}} for {{Autonomous Driving}},'' in \emph{Intelligent Vehicles Symposium (IV)}, 2022.

\bibitem{zhou_data}
J.~Zhou and J.~Beyerer, ``Corner cases in data-driven automated driving: Definitions, properties and solutions,'' in \emph{Intelligent Vehicles Symposium (IV)}, 2023.

\bibitem{heideckerCornerCaseDefinition}
F.~Heidecker, M.~Bieshaar, and B.~Sick, ``Corner {{Case Definition}} in {{Machine Learning Processes}} for the {{Perception}} of {{Highly Automated Driving}},'' \emph{AI Perspectives \& Advances}, vol.~6, 2024.

\bibitem{breitensteinSystematizationCornerCases2020}
J.~Breitenstein, J.~A. Termohlen, D.~Lipinski, and T.~Fingscheidt, ``Systematization of {{Corner Cases}} for {{Visual Perception}} in {{Automated Driving}},'' in \emph{Intelligent Vehicles Symposium (IV)}, 2020.

\bibitem{breitenstein_corner_2021}
J.~Breitenstein, J.-A. Term{\"o}hlen, D.~Lipinski, and T.~Fingscheidt, ``Corner cases for visual perception in automated driving: Some guidance on detection approaches,'' \emph{arXiv preprint:2102.05897}, 2021.

\bibitem{bogdollDescriptionCornerCases2021}
D.~Bogdoll, J.~Breitenstein, F.~Heidecker, M.~Bieshaar \emph{et~al.}, ``Description of {{Corner Cases}} in {{Automated Driving}}: {{Goals}} and {{Challenges}},'' in \emph{International Conference on Computer Vision (ICCV) Workshops}, 2021.

\bibitem{bogdollAnomalyDetectionAutonomous2022}
D.~Bogdoll, M.~Nitsche, and J.~M. Zollner, ``Anomaly {{Detection}} in {{Autonomous Driving}}: {{A Survey}},'' in \emph{Conference on Computer Vision and Pattern Recognition (CVPR) Workshops}, 2022.

\bibitem{vaterSystematicApproachDefinition2023}
L.~Vater, M.~Sonntag, J.~Hiller, P.~Schaudt, and L.~Eckstein, ``A {{Systematic Approach Towards}} the {{Definition}} of the {{Terms Edge Case}} and {{Corner Case}} for {{Automated Driving}},'' in \emph{International Conference on Electrical, Computer, Communications and Mechatronics Engineering (ICECCME)}, 2023.

\bibitem{liuCurseRarityAutonomous2024}
H.~X. Liu and S.~Feng, ``Curse of rarity for autonomous vehicles,'' \emph{Nature Communications}, vol.~15, no.~1, 2024.

\bibitem{li_coda_2022}
K.~Li, K.~Chen, H.~Wang, L.~Hong \emph{et~al.}, ``{CODA}: {A} {Real}-{World} {Road} {Corner} {Case} {Dataset} for {Object} {Detection} in {Autonomous} {Driving},'' in \emph{European Conference on Computer Vision (ECCV)}, 2022.

\bibitem{bogdoll_anomaly_2022}
D.~Bogdoll, M.~Nitsche, and J.~M. Z\"{o}llner, ``{Anomaly Detection in Autonomous Driving: A Survey},'' in \emph{Conference on Computer Vision and Pattern Recognition (CVPR) Workshops}, 2022.

\bibitem{wong_identifying_2020}
K.~Wong, S.~Wang, M.~Ren, M.~Liang, and R.~Urtasun, ``Identifying {Unknown} {Instances} for {Autonomous} {Driving},'' in \emph{Conference on Robot Learning {CORL}}, 2020.

\bibitem{nunes_unsupervised_2022}
L.~Nunes, X.~Chen, R.~Marcuzzi, A.~Osep \emph{et~al.}, ``Unsupervised class-agnostic instance segmentation of 3d lidar data for autonomous vehicles,'' \emph{IEEE Robotics and Automation Letters}, vol.~7, no.~4, 2022.

\bibitem{cen_open-set_2021}
J.~Cen, P.~Yun, J.~Cai, M.~Wang, and M.~Liu, ``Open-set 3d object detection,'' \emph{International Conference on 3D Vision (3DV)}, 2021.

\bibitem{bogdoll_perception}
D.~Bogdoll, S.~Uhlemeyer, K.~Kowol, and J.~M. Z{\"o}llner, ``Perception datasets for anomaly detection in autonomous driving: A survey,'' in \emph{Intelligent Vehicles Symposium (IV)}, 2023.

\bibitem{geiger_are_2012}
A.~Geiger, P.~Lenz, and R.~Urtasun, ``Are we ready for autonomous driving? {The} {KITTI} vision benchmark suite,'' in \emph{Conference on Computer Vision and Pattern Recognition (CVPR)}, 2012.

\bibitem{Geppert_Anomaly_2023_BA}
V.~Geppert, ``{Anomaly Detection with Model Contradictions for Autonomous Driving},'' Bachelor Thesis, {Karlsruhe Institute of Technology (KIT)}, 2023.

\bibitem{niu2024unsupervised}
D.~Niu, X.~Wang, X.~Han, L.~Lian, R.~Herzig, and T.~Darrell, ``Unsupervised universal image segmentation,'' in \emph{Conference on Computer Vision and Pattern Recognition (CVPR)}, 2024.

\bibitem{nie2024lanecorrectselfsupervisedlanedetection}
M.~Nie, X.~Cai, H.~Xu, and L.~Zhang, ``Lanecorrect: Self-supervised lane detection,'' \emph{arXiv: 2404.14671}, 2024.

\bibitem{mayr_ssl_drivable}
J.~Mayr, C.~Unger, and F.~Tombari, ``Self-supervised learning of the drivable area for autonomous vehicles,'' in \emph{International Conference on Intelligent Robots and Systems (IROS)}, 2018.

\end{thebibliography}

\end{document}